\documentclass[11pt]{article}
\usepackage{coling2020}
\usepackage{times}
\usepackage{url}
\usepackage{latexsym}
\usepackage{graphicx}
\usepackage{multibib} 
\usepackage{multirow}
\usepackage{amssymb}
\usepackage{pifont}
\usepackage{amsmath}
\usepackage{bm}
\usepackage[bottom]{footmisc}
\usepackage{CJKutf8}

\usepackage{coling2020}
\usepackage{times}
\usepackage{url}
\usepackage{latexsym}
\usepackage{pgfplots}
\usepackage{caption}
\usepackage{graphicx}
\usepackage{hyperref}

\colingfinalcopy 

\title{What are We Depressed about When We Talk about COVID19: \\Mental Health Analysis on Tweets Using Natural Language Processing}

\author{Irene Li$^{1,*}$, Yixin Li$^1$, Tianxiao Li$^1$, \\ 
\textbf{Sergio Alvarez-Napagao}$^2$ \textbf{,} \textbf{Dario Garcia-Gasulla}$^2$ \textbf{,} \and \textbf{Toyotaro Suzumura}$^2$ \\
$^1$Yale University, USA \\
$^2$Barcelona Supercomputing Center (BSC), Spain
}

\date{}

\begin{document}
\maketitle
\begin{abstract}
  The outbreak of coronavirus disease 2019 (COVID-19) recently has affected human life to a great extent.  Besides direct physical and economic threats, the pandemic also indirectly impact people’s mental health conditions, which can be overwhelming but difficult to measure.  The problem may come from various reasons such as unemployment status, stay-at-home policy, fear for the virus, and so forth. In this work, we focus on applying natural language processing (NLP) techniques to analyze tweets in terms of mental health. We trained deep models that classify each tweet into the following emotions: anger, anticipation, disgust, fear, joy, sadness, surprise and trust. We build the EmoCT (Emotion-Covid19-Tweet) dataset for the training purpose by manually labeling 1,000 English tweets. Furthermore, we propose and compare two methods to find out the reasons that are causing sadness and fear. 
\end{abstract}

\section{Introduction}
Mental health is becoming a common issue. According to World Health Organization (WHO), one in four people in the world will be affected by mental or neurological disorders at some point in their lives\footnote{\url{https://www.who.int/whr/2001/media\_centre/press\_release/en/}}. 
A large emergency, such as the coronavirus disease 2019 (COVID-19), would especially sharply increase people's mental health problems, not only from the emergency itself, but also from the subsequent social outcomes such as unemployment, shortage of resources and financial crisis. Almost all people affected by emergencies will experience psychological distress, which for most people will improve over time\footnote{\url{https://www.who.int/news-room/fact-sheets/detail/mental-health-in-emergencies}}. In order to help the society get prepared in response to surging mental problems during and after COVID-19 emergency, we need to understand people's general mental status as a first step. 

Language, as a direct tool for people to convey their feelings and emotions, can be very useful and helpful in the estimation of mental health conditions. Nowadays, people post their thoughts and experiences on social media including Facebook, Instagram, and Twitter. Especially, due to the recent impact of COVID-19, a large number of people move their works online, making some users are even more active than usual. Previous works have been conducted to utilize natural language processing (NLP) methods to process internet-based text data such as posts, tweets, and text messages on mental health problems \cite{althoff2016large,calvo2017natural,larsen2015we,dini2016emotion}.

There are mainly three challenges in working with tweets using NLP methods. The first challenge is the large number of new posts online but restricted availability of APIs. There may be up to 90 or even 100 million tweets per day \cite{calvo2017natural}, so most of research is conducted on random samples \cite{ritter2011named,mohammad2017stance,pandey2017twitter}. We are interested in a million-level of tweets and also in a larger time span.
Another challenge is that many existing research only focused on English tweets \cite{Farruque2019BasicAD,dini2016emotion}. The \textit{We Feel} platform by \newcite{larsen2015we} deals with real-time tweets in a large-scale but only can process English ones. To understand the global influence of corona virus, and estimate the emotion variation across culture and region, we want to utilize texts in multiple languages. The third challenge is the lack of labeled dataset for COVID-19. Though there exist labeled Twitter dataset for sentiment and emotions \cite{go2009twitter,mohammad2017stance,hasan2014emotex}, due to the domain discrepancy, we still wish to have a manually-labeled dataset for training to have a better performed model.

The work by \newcite{larsen2015we} applies principal component analysis (PCA) to predict emotions. \newcite{abidin2017n} proposed to use k-Nearest Neighbors and Naive Bayes classifier to do classification on tweets. A recent work by \newcite{Farruque2019BasicAD} applied deep models to do multi-label classification on tweets. Very recently, many types of contextualized word embeddings are proposed and substantially improved the performance on many NLP tasks. A new language representation model, BERT \cite{devlin2018bert}, was proposed and obtains competitive results on up to 11 NLP tasks including classification, natural language inference and question answering. 
In this work, we apply a pre-trained BERT and fine-tune on our labeled data, providing in-depth analysis of mental health. 

Our contributions are three-fold: we build the \textbf{EmoCT} (\textbf{Emo}tion-\textbf{C}OVID19-\textbf{T}weet) dataset for classifying COVID-19-related tweets into eight emotions; then, we propose two models to do both single-label and multi-label classification respectively based on a multilingual BERT model, which are capable to predict on up to 104 languages and achieving promising results on English tweets; further analysis on case studies provide clues to understand why and how the public may feel fear and sad about COVID-19.


\section{Dataset}
\begin{figure}[h]
\centering
\includegraphics[width=16cm]{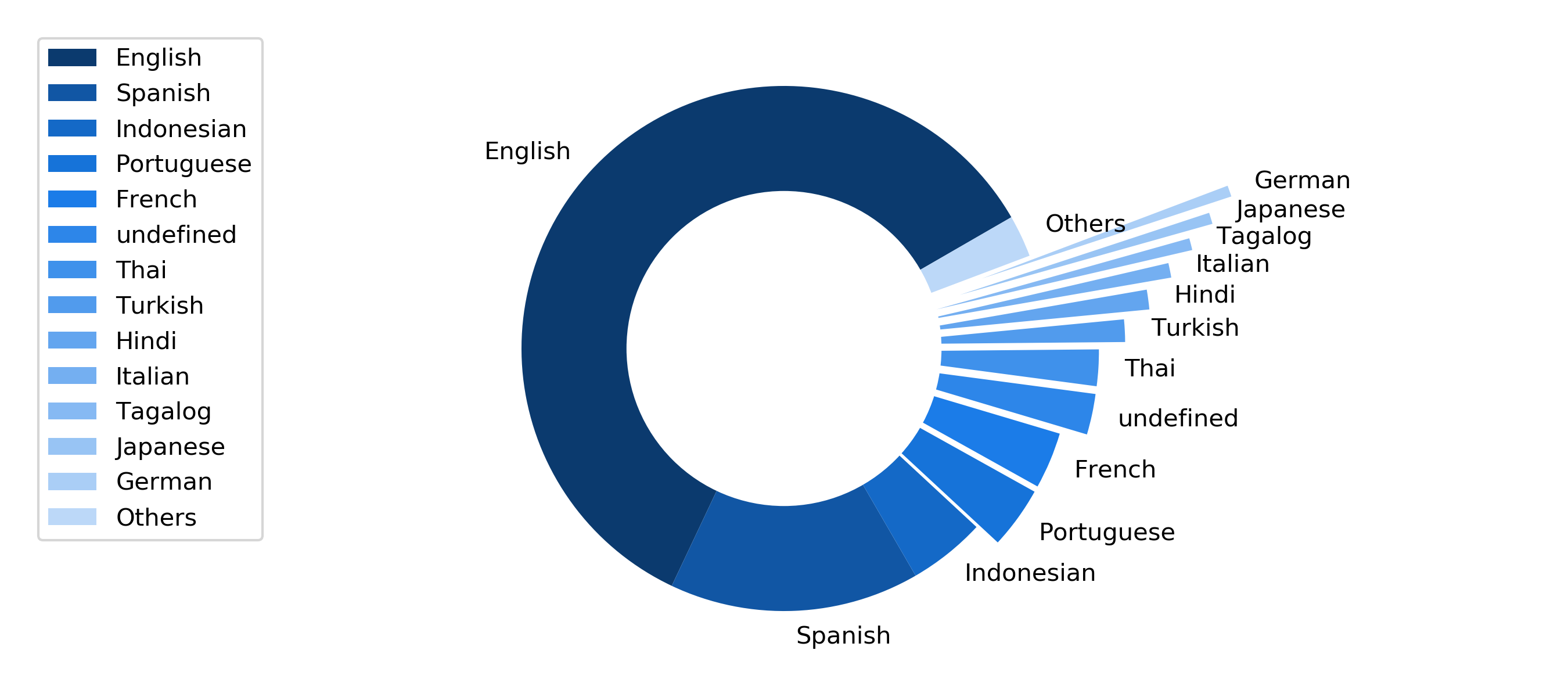}
\caption{Language distribution on 8,148,202 tweets. }
\label{fig:language}
\end{figure}

\begin{figure}[h]
\centering
\includegraphics[width=14cm]{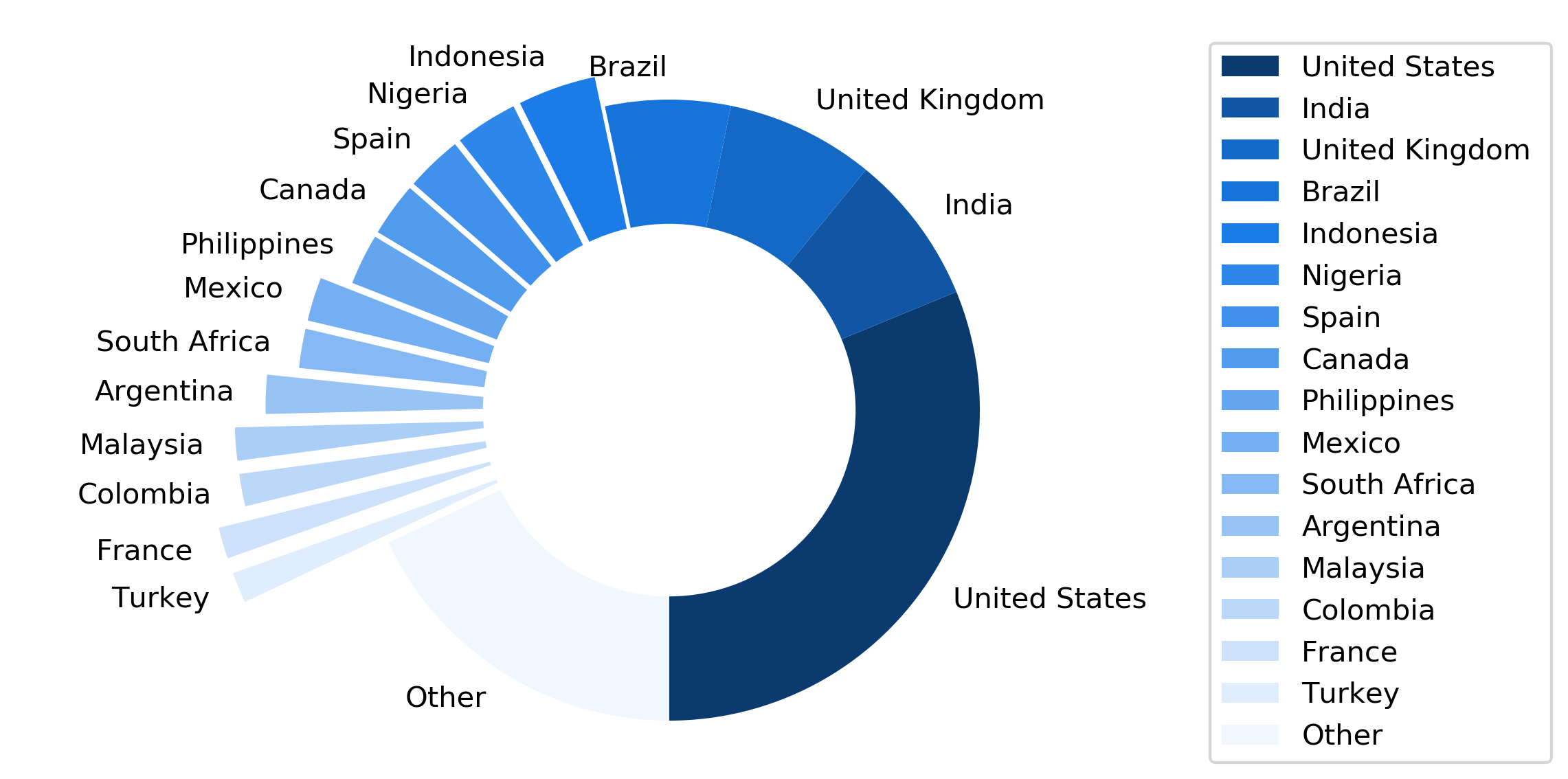}
\caption{Geolocation distribution on 76,460 tweets. }
\label{fig:geo}
\end{figure}

We applied Twitter API\footnote{\url{https://developer.twitter.com/en/docs/tweets/data-dictionary/overview/intro-to-tweet-json}} to conduct a crawler with a list of keywords:coronavirus, covid19, covid, COVID-19, covid\_19, confinamiento, flu, virus, hantavirus, fever, cough, social distance, lockdown, pandemic, epidemic, conlabelious, infection, stayhome, corona, épidémie, epidemie, epidemia, 
\begin{CJK*}{UTF8}{gbsn}新冠肺炎\end{CJK*},
 \begin{CJK}{UTF8}{gbsn}新型冠狀病毒\end{CJK},
 \begin{CJK}{UTF8}{gbsn}疫情\end{CJK},
\begin{CJK}{UTF8}{gbsn}新冠病毒\end{CJK}, 
\begin{CJK}{UTF8}{gbsn}感染\end{CJK},
\begin{CJK}{UTF8}{min}新型コロナウイルス\end{CJK},
\begin{CJK}{UTF8}{min}コロナ\end{CJK}. 
Each day, we are able to crawl 3 million tweets in free text format from different languages. Due to the high capacity, we look at the tweets from March 24 to 26, 2020 to get language and geolocation statistics. Among these tweets, 8,148,202 tweets have the language information (\texttt{lang} field of the \texttt{Tweet} Object in Tweet API), and 76,460 tweets have the geographic information (\texttt{country\_code} value from the \texttt{place} field if not none). We show the distributions in Figure~\ref{fig:language} and \ref{fig:geo}. 

To train the models for classification, we built \textbf{EmoCT} (\textbf{Emo}tion-\textbf{C}ovid19-\textbf{T}weet) dataset. We randomly annotated 1,000 English tweets selected from our crawled data. Following the work of EmoLex \cite{mohammad2013crowdsourcing,hasan2014emotex}, we classify each tweet into the following emotions: anger, anticipation, disgust, fear, joy, sadness, surprise and trust. Each tweet is labeled as one, two or three emotion labels. For each emotion, we made sure that the primary label appeared in 125 tweets, and there is no number control in the secondary and tertiary label. We then split into 100/25 for each emotion as the training/testing set. We release two versions of the dataset: single-labeled version where only the primary label is kept for each example, and multi-labeled version where all the labels are kept. In this way, both single-label classification and multi-label classification can be conducted. We release the EmoCt dataset to the public \footnote{\url{https://github.com/IreneZihuiLi/EmoCT}}, where only Tweet IDs and labels can be found by eliminating the actual texts due to corresponding restrictions.

\section{Classification}

\textbf{Single-label Classification} We first attempt to do a single-label classification task based on the single-labeled version of EmoCT. We apply a pre-trained multilingual version BERT model\footnote{https://github.com/huggingface/transformers: \textit{bert-base-multilingual-cased} model}. We take the output of the \texttt{[CLS]} token and add a fully-connected layer, which is fine-tuned using the labeled training examples (BERT). We set the learning rate to be $10^{-5}$ and number of epochs to be 20. Besides, we also fine-tune with the MLM (masked language model) on 1,181,342 unlabeled tweets randomly selected from our crawled data, and then trained on EmoCT (BERT(ft)). Table~\ref{tab:single} shows the performance of the two models. As we can see, both models have competitive results on accuracy and F1, and BERT(ft) performs slightly better than BERT, so we take this model as our main model for analysis in later sections.

\begin{table}[h!]
\centering
\begin{tabular}{ccc}\hline 
Method       & Accuracy & F1    \\\hline \hline
BERT & 0.9549 & 0.9545 \\
BERT(ft) & 0.9562 & 0.9558 \\ \hline
\end{tabular}
\caption{Single-label Classification Results on EmoCT single-labeled version.}
\label{tab:single}
\end{table}


%

\begin{table}[b]
\centering
\begin{tabular}{cccc}\hline \hline 
Method   & Average precision & Coverage error & Ranking loss    \\\hline 
BERT & 0.6415 & 3.2261 & 0.2325 \\
BERT(ft) & 0.6467 & 3.1256 & 0.2159\\ \hline\hline 
\end{tabular}
\caption{Multi-label Classification Results on EmoCT multi-labeled version.}
\label{tab:multi}
\end{table}

\textbf{Multi-label Classification} We also perform multi-label classification on the multi-labeled version of EmoCT. In this setting, each tweet has up to three labels out of eight, and we assume the labels are independent. We build a single-layer classifier with the activation function to be Sigmoid, which receives BERT output and predicts the possibility of containing each of the eight labels (BERT). The model uses binary cross-entropy loss and is trained for 10 epochs with learning rate $10^{-5}$. Similarly, we also compare with a fine-tuned version as did in the previous model (BERT(ft)). For evaluation, we use example-based evaluation metrics mentioned in the work of \newcite{zhang2014multiclass} in Table~\ref{tab:multi}. We could see that the two models achieve relatively low scores, probably due to the small-scale training data. In Table~\ref{tab:multiROC}, we show the area-under-curve (AUC) of the response operating characteristic (ROC) curve for each class and their micro average. It can be noticed that both models are not performing so well by looking at the average score, and they are not very confident on certain classes like \textit{anticipation}, and we leave it as future work.

\begin{table}[t!]
\centering
\small
\begin{tabular}{cccccccccc}\hline \hline 
Method   & Anger & Anticipation & Disgust   &  Fear &  Joy &    Sadness & Surprise & Trust & Micro-Avg.\\\hline 
BERT & 0.7473 & 0.6173 & 0.8222 & 0.7010 & 0.8380  & 0.7394 & 0.8620 & 0.7919 & 0.7778  \\
BERT(ft) & 0.7397 & 0.6897 & 0.8364 & 0.7344 & 0.8430  & 0.6809 & 0.8676 & 0.8228 & 0.7891 \\ \hline \hline 
\end{tabular}
\caption{AUROC for each label of multi-label Classification on EmoCT multi-labeled version, as well as the micro average over all classes.}
\label{tab:multiROC}
\end{table}

\begin{figure}[h]
\centering
\includegraphics[width=0.6\textwidth]{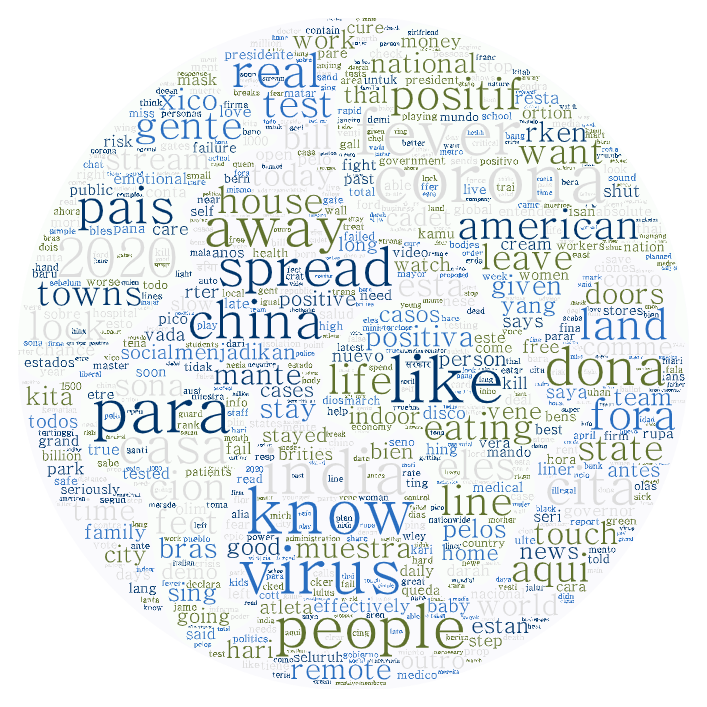}
\caption{Wordcloud from attention.}
\label{fig:att}
\end{figure}

\section{Correlation}
Due to the outbreak of coronavirus emergency, the two emotions \textit{sad} and \textit{fear} are more related to severe negative sentiments like depressed. To understand why the public may feel fear and sadness, we then attempt to analyze words and phrases that have a high correlation with both emotions. We apply our BERT(ft) model from the single-label classification task to predict the emotion label on randomly-picked 1 million tweets data on April 7, 2020. Then we compare two methods to do further analysis. Note that we keep only the tweets labeled as fear and sadness.

\textbf{Attention Weight} When predicting the emotion label for each tweet, we take the last attention layer of the model and collect the top 3 tokens which have the maximum attention weights. Finally we rank the tokens by frequency and plot the wordcloud\footnote{Visualization tool: https://wordart.com/. Invalid for a few languages.} of the top 500 tokens after filtering some stopwords in Figure~\ref{fig:att}. A drawback of this method is that the tokens are split, so we can see some keywords that may not be meaningful without contexts, for example: \textit{like}, \textit{know} and \textit{2020}. However, we can get some reasonable keywords: \textit{fever}, \textit{corona}, \textit{spread}, \textit{virus} and so on. Such words appear with a high frequency in the tweets labeled as fear and sadness, which may explain what and why people are feeling fear or sad. Note that this method can handle multiple language input as the pre-trained BERT model supports 104 different languages, though training was conducted on an English corpus.

\begin{figure}[t]
\centering
\includegraphics[width=0.6\textwidth]{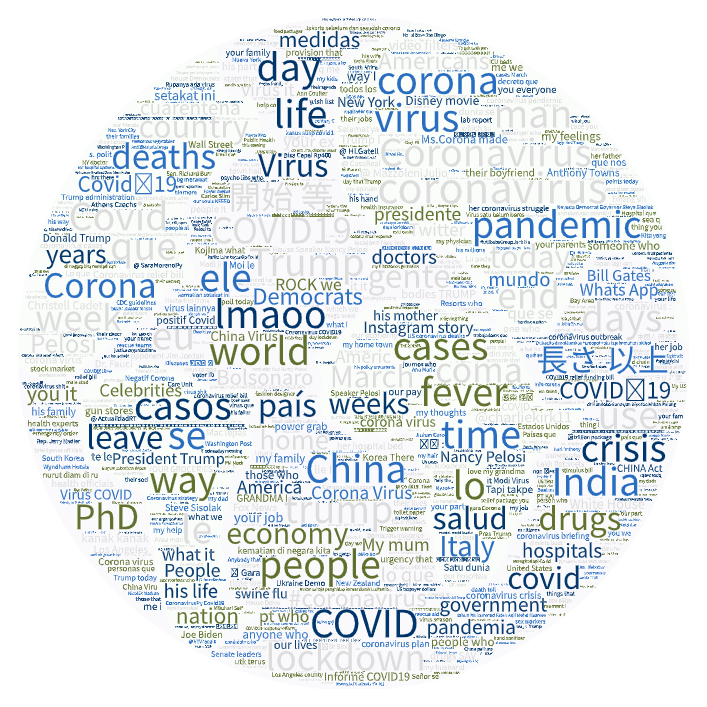}
\caption{Wordcloud from POS labelging.}
\label{fig:pos}
\end{figure}

\begin{figure}[b]
\centering
\includegraphics[width=0.9\textwidth]{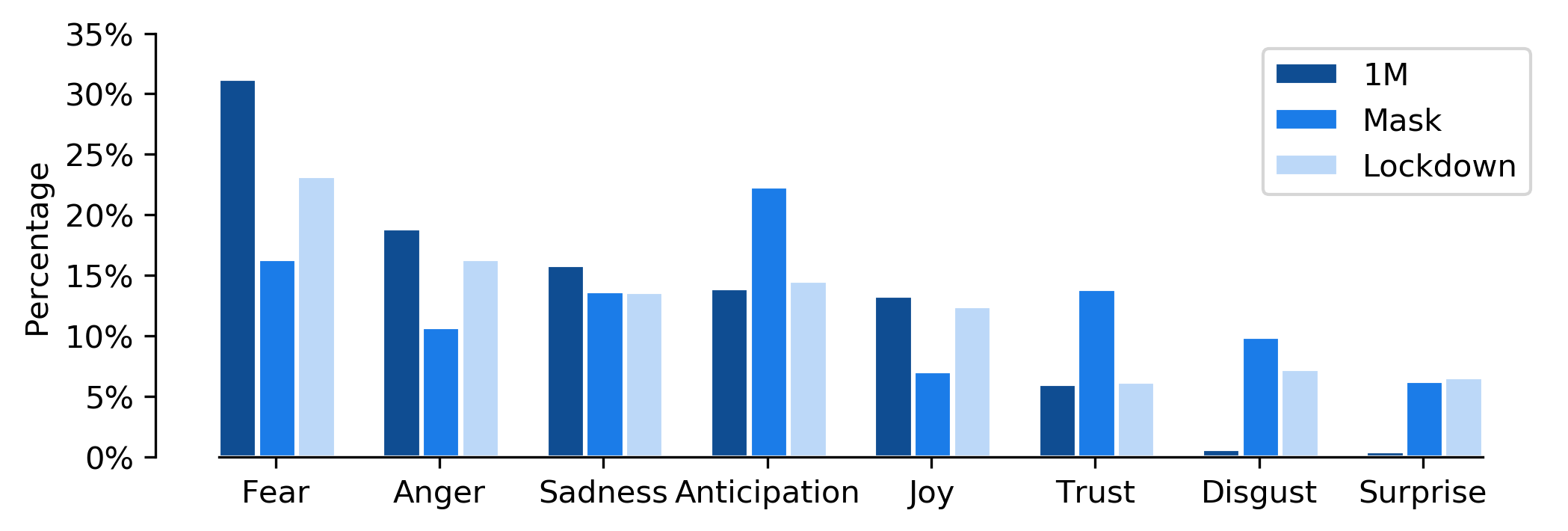}
\caption{Comparison of emotion distribution.}
\label{fig:comp}
\end{figure}

\textbf{POS tagging} Intuitively, we assume that nouns are more meaningful in a tweet, making it possible and easier to understand the reasons why it is labeled as fear or sadness. As a comparison, we look at the Part-of-Speech (POS) tag of each token in the tweets and keep the nouns and noun phrases only. We apply the Stanza Python library to do POS tagging \cite{qi2020stanza} and we include supporting to six languages including English, Spanish, Portuguese, Japanese, German and Chinese. Similarly, we plot the top 500 keywords and phrases based on frequency in Figure ~\ref{fig:pos}. There are some informative keywords and phrases captured: \textit{pandemic}, \textit{China}, \textit{economy}, \begin{CJK}{UTF8}{min}開始\end{CJK} (means \textit{starting} in English), \textit{President Trump}, \textit{White House} and so on. While working on the analysis, we saw other meaningful phrases such as \textit{gun stores}, \textit{school closings}, and \textit{health conditions} which has a lower frequency and may not be visible.

\section{Emotion Trend Analysis}

The emotion trend among different hashtags or topics is also very important, as it potentially may show the public attitude change within a period of time. We still choose the single-label classification BERT(ft) model to do prediction. We provide a case study on two words: \textit{mask} and \textit{lockdown}. We first pick 1 million tweets randomly from the data of the date March 29th, 2020. By filtering on the keywords, we found 8,071 tweets that contain the word \textit{mask}, and 31,146 tweets that contain the word \textit{lockdown}. Figure \ref{fig:comp} shows the comparison of emotion distribution among 1 million samples (1M), tweets with \textit{mask}, and tweets with \textit{lockdown}. In the 1 million data, most tweets are classified into negative classes like fear, anger and sadness. But when people are talking about masks, more tweets are classified into anticipation and trust, which is sometimes more neutral and positive. For the tweets talking about lockdown, there is no significant difference with that of 1M.

\begin{figure}[!h]
\centering
\includegraphics[width=0.9\textwidth]{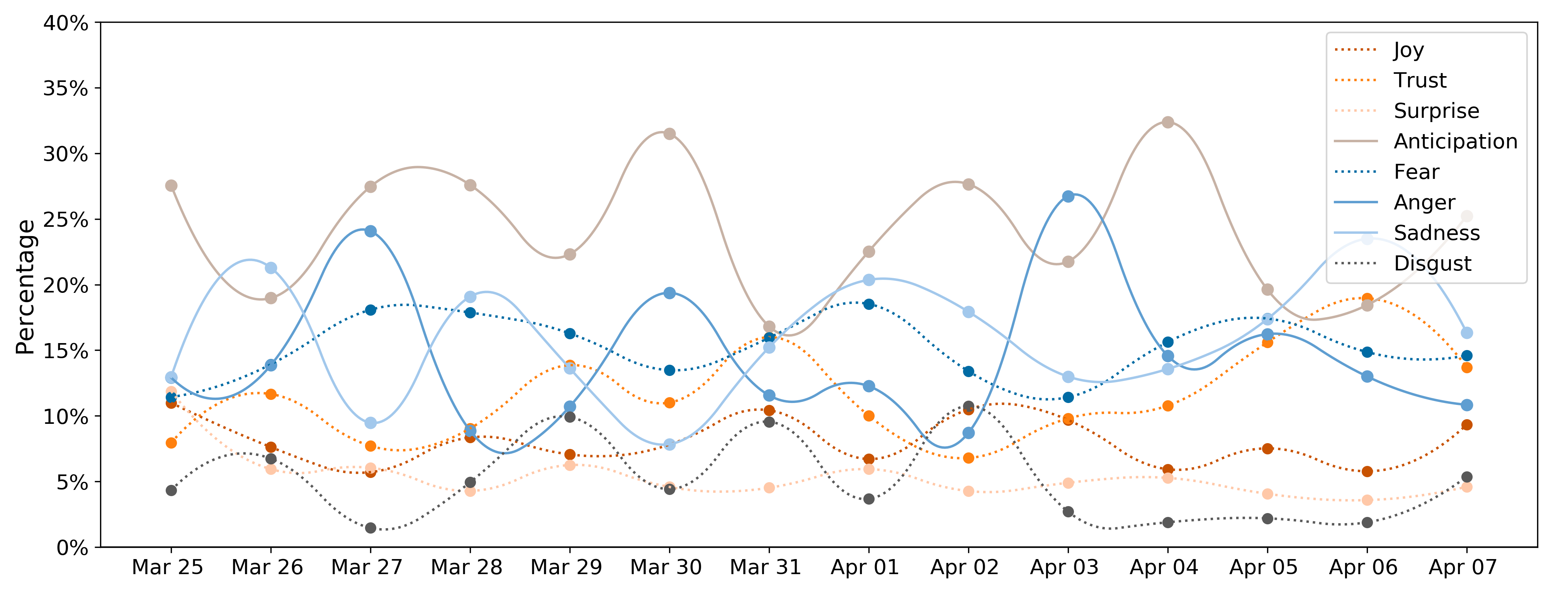}
\caption{Emotion trend on the word \textit{mask} from March 25  2020 to April 7, 2020.}
\label{fig:mask}
\end{figure}

\begin{figure}[!h]
\centering
\includegraphics[width=0.9\textwidth]{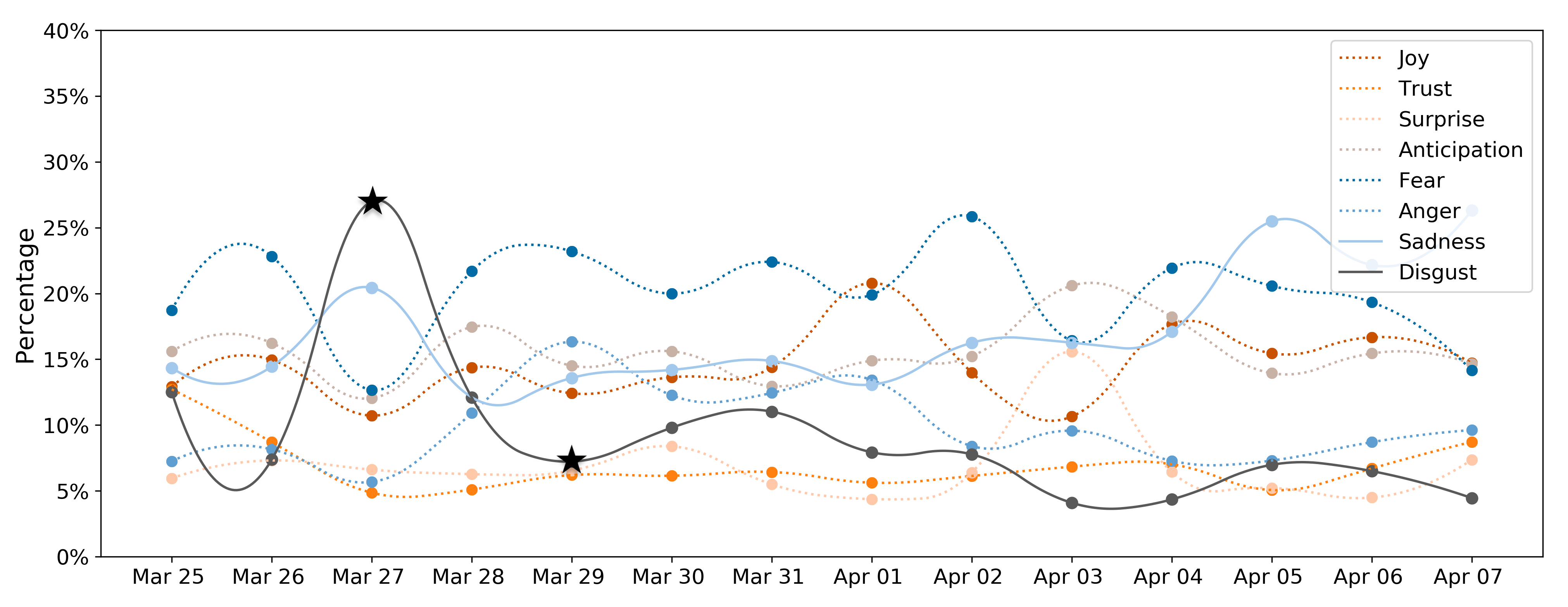}
\caption{Emotion trend on the word \textit{lockdown} from March 25  2020 to April 7, 2020.}
\label{fig:lockdown}
\end{figure}

To further analyze the trends, we select the data of two weeks (March 25, 2020-April 7, 2020), and apply the same model to predict the emotion labels on all the tweets we crawled (around 3 million each day) that contains the two mentioned keywords respectively. There is no significant change for the emotion distribution in all the data. However, we found the dominating emotions and variations of the change are closely related to the topic. In Figure \ref{fig:mask} and \ref{fig:lockdown}, we illustrate the emotion trend for each single day of the selected keywords. The high variation (plot in solid lines in the figures) showed up in \textit{sadness}, \textit{anger} and \textit{anticipation} for the tweets that contain the word \textit{mask} in Figure \ref{fig:mask}, and \textit{disgust}, \textit{sadness} for the tweets that contain the word \textit{lockdown} in Figure \ref{fig:lockdown}. Especially, for the \textit{lockdown} tweets, the percentage of \textit{disgust} emotion had a significant increase on March 27 and dropped on the next two days, as marked with the black asterisks. 
To further investigate, we looked at the news in March 27, which included U.S. as the first country to report 100,000 confirmed coronavirus cases, and 9 in 10 Americans were staying home; India and South Africa joined the countries to impose lockdowns. Given that the United States, India and Brazil have large group of twitter users, we assume that this dramatic change may be triggered by those news.

\section{Conclusion and Future Work}
In this work, we build the EmoCT dataset for classifying COVID-19-related tweets into different emotions. Based on this dataset, we conducted both single-label and multi-label classification tasks and achieved promising results. Besides, to understand the reasons why the public may feel sad or fear, we applied two methods to calculate correlations of the keywords. 

In the future work, we will study more in-depth analysis to better understand how COVID-19 affect on mental health. It is possible to have detailed statistics and analysis grouped by languages and locations. In addition, we are planning to collect a multi-lingual version of the existing EmoCT dataset to promote related research.

With the capability of tracking twitter data in a longer term, we want to investigate how people recover from this global COVID-19 crisis from sadness and fear, and rebuild trust and joy to the society. We are interested to understand the relationship of mental health curve and COVID-19 case/mortality rate curve, the variation of emotion changes among different region and culture. It will be helpful for us to have a correct estimates of the COVID-19 effects on people's long term mental health, and be prepared for the next crisis. Besides, it is also possible to crawl the tweets before the outbreak of COVID-19 and study how the mental health related issues are changed between, before and after COVID-19.  We present more details in our website \url{https://www.covid19analytics.org/}.



\bibliographystyle{coling}
\bibliography{coling2020}

\end{document}